\documentclass[conference]{IEEEtran}
\IEEEoverridecommandlockouts
\usepackage{algorithmic}
\usepackage{graphicx}
\usepackage{textcomp}
\usepackage{times}
\usepackage{epsfig}
\usepackage{enumitem}
\usepackage[space]{cite}
\usepackage{color}
\usepackage{amsmath,amssymb,amsopn,amstext,amsfonts}
\usepackage{mathtools}
\usepackage{indentfirst}
\usepackage{subfigure}
\usepackage{multirow}
\usepackage{makecell}
\usepackage{comment}
\usepackage{etoolbox}
\usepackage{url}
\usepackage{caption}
\graphicspath{ {./figures/} }

\usepackage{balance}

\usepackage{array}
\newcolumntype{L}[1]{>{\raggedright\let\newline\\\arraybackslash\hspace{0pt}}m{#1}}
\newcolumntype{C}[1]{>{\centering\let\newline\\\arraybackslash\hspace{0pt}}m{#1}}
\newcolumntype{R}[1]{>{\raggedleft\let\newline\\\arraybackslash\hspace{0pt}}m{#1}}

\DeclareGraphicsExtensions{.pdf,.png,.jpg,.eps,.PNG}

\begin{document}

\title{FP-Stereo: Hardware-Efficient Stereo Vision for Embedded Applications}

\author{\IEEEauthorblockN{Jieru Zhao$^{1}$, Tingyuan Liang$^1$, Liang Feng$^2$, Wenchao Ding$^1$, Sharad Sinha$^3$, Wei Zhang$^1$ and Shaojie Shen$^1$}
 \IEEEauthorblockA{\textit{$^1$Hong Kong University of Science and Technology}, \textit{$^2$Alibaba Group,}
 \textit{$^3$India Institute of Technology Goa}\\
 \{jzhaoao, tliang, lfengad, wdingae, wei.zhang, eeshaojie\}@ust.hk, sharad\_sinha@ieee.org}
}

\maketitle

\begin{abstract}
Fast and accurate depth estimation, or stereo matching, is essential in embedded stereo vision systems, requiring substantial design effort to achieve an appropriate balance among \textit{accuracy}, \textit{speed} and \textit{hardware cost}. To reduce the design effort and achieve the right balance, we propose FP-Stereo for building high-performance stereo matching pipelines on FPGAs automatically. FP-Stereo consists of an open-source hardware-efficient library, allowing designers to obtain the desired implementation instantly. Diverse methods are supported in our library for each stage of the stereo matching pipeline and a series of techniques are developed to exploit the parallelism and reduce the resource overhead. To improve the usability, FP-Stereo can generate synthesizable C code of the FPGA accelerator with our optimized HLS templates automatically. To guide users for the right design choice meeting specific application requirements, detailed comparisons are performed on various configurations of our library to investigate the accuracy/speed/cost trade-off. Experimental results also show that FP-Stereo outperforms the state-of-the-art FPGA design from all aspects, including 6.08\% lower error, 2x faster speed, 30\% less resource usage and 40\% less energy consumption. Compared to GPU designs, FP-Stereo achieves the same accuracy at a competitive speed while consuming much less energy.
\end{abstract}
\section{Introduction}
Real-time and robust stereo vision systems for the computation of depth information are increasingly popular in many embedded applications including robotic navigation\cite{huntsberger2011stereo,gao2020autonomous} and autonomous vehicles\cite{Menze2018JPRS,zhang2020efficient}. Stereo matching methods take a pair of left-right images from stereo cameras as input and generate the disparity map for depth estimation. Typical stereo matching algorithms can be classified into two categories: local and global methods. Local methods aggregate matching costs over a local region, achieving fast speeds but suffering from textureless and discontinuous regions\cite{tao2008fast,yoon2006adaptive}. Global methods estimate the disparity map by minimizing a global energy function, using graph cuts\cite{kolmogorov2001computing} or belief propagation\cite{sun2003stereo,klaus2006segment}, which achieve high accuracy but are computationally inefficient. Semi-global matching (SGM) approximates the global optimization by minimizing a pathwise energy function and aggregating matching costs in multiple directions instead of the whole image\cite{hirschmuller2007stereo}. Due to the excellent trade-off between accuracy and speed, SGM has become one of the most widely used stereo matching algorithms, especially in real-world embedded applications. \\
\indent Most existing SGM designs on FPGAs were implemented at register-transfer level (RTL)\cite{wang2015real,li2017high,roszkowski2014fpga,banz2010real,gehrig2009real,buder2012dense,hofmann2016scalable}, and are hard for reproduction and modification. Relying on high-level synthesis (HLS), which automatically synthesizes C code into RTL designs, the development time can be shortened significantly. Several acceleration approaches \cite{rupnow2011high,liang2012high,rahnama2018real,ali2017exploring} utilize HLS to map \textit{local} stereo matching algorithms on FPGAs, achieving fast speed while incurring considerable accuracy loss. By comparison, SGM is more practical with higher accuracy. However, extra design effort is necessary to achieve real-time speeds on specialized hardware due to the huge pressure on hardware resources caused by intermediate computing results\cite{wang2015real}. Rahnama et al. \cite{rahnama2018r3sgm} implement an SGM variation on FPGA using HLS with a throughput of 72 frames per second (FPS) for 1242*375 image size and 128 disparity levels on the KITTI dataset \cite{menze2015object}. Recently, Xilinx released xfOpenCV \cite{xilinxxfopencv}, and its SGM implementation achieves a faster speed of 81 FPS on KITTI images. Although raising the abstraction level, neither method \cite{rahnama2018r3sgm,xilinxxfopencv} can fully exploit the parallelism to achieve competitive speeds compared to RTL designs. Moreover, apart from \textit{efficiency} concerns, high-quality stereo matching pipelines should achieve user-required \textit{accuracy} given limited \textit{hardware budgets}. However, the implementations in \cite{rahnama2018r3sgm} and \cite{xilinxxfopencv} only support a single method for each stage of the SGM pipeline, lacking the flexibility to adapt to various user requirements in \textit{accuracy}, \textit{speed} and \textit{hardware cost}.\\
\begin{figure*}[t]
  \centering
  \includegraphics[width=0.98\textwidth]{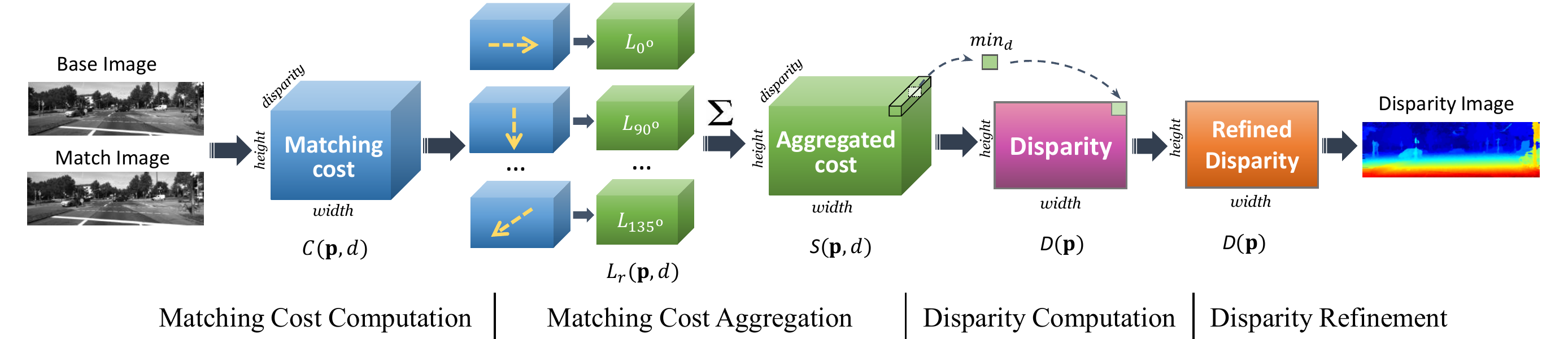}
  \caption{Diagram of the semi-global matching algorithm.}
  \label{SGM flow}
\end{figure*}
\indent To this end, we propose FP-Stereo for building high-performance SGM pipelines on FPGAs automatically to meet different accuracy/speed/cost requirements. Our contributions are summarized as follows:
\begin{itemize} [leftmargin=0.5cm, topsep=0pt,itemsep=-1ex,partopsep=1ex,parsep=1.2ex]
\item We provide an open-source hardware-efficient library on FPGA, composed of optimized HLS C kernels, for embedded stereo vision applications.
\item We apply effective optimization and implementation techniques to our library templates, fully exploiting the parallelism and greatly reducing the resource overhead.
\item We present several methods to improve usability and scalability of our library, including auto-computed data widths, user-friendly interface and automatic code generation.
\item We investigate how algorithmic choices, i.e., various functions or parameters, impact hardware performance, and how hardware budgets constrain algorithmic choices. Our findings can serve as guidance for users to select the right design choice which achieves an appropriate accuracy/speed/resource balance on FPGA.
\item FP-Stereo supports high-quality SGM implementations. Compared to xfOpenCV, FP-Stereo achieves higher accuracy (6.08\% lower error) at 2x speed (161 FPS) while consuming 30\% fewer resources and 40\% less energy at full KITTI resolution. Compared to GPU designs, FP-Stereo achieves the best speed/power ratio for the same accuracy.
\end{itemize}

\section{Pipeline of Semi-Global Matching Algorithm}\label{sgm methods}
Figure \ref{SGM flow} illustrates a high-level overview of the semi-global matching algorithm, which contains four stages: 
(1) similarity comparison is applied on the base (left) and match (right) images to compute the matching cost for each pixel at each disparity, generating a matching cost volume; (2) to smooth the cost volume, matching costs of neighboring pixels at each disparity are aggregated in different directions; (3) the disparity map of the base image is computed from the aggregated cost volume; and (4) several refinement steps are applied to remove outliers. 
In this section, we will introduce frequently used methods for each stage of the stereo matching algorithm. All of them are supported by FP-Stereo.
\subsection{Matching Cost Computation}\label{cost functions}
Matching costs measure the similarity of corresponding pixels\cite{hirschmuller2008evaluation}.
Given pixel $\textbf{p}=(x,y)$ in the base image, the corresponding pixel at disparity \textit{d} in the rectified match image is \textbf{p}-\textit{d}, where $d=\{0,1,2,...,d_\textit{max}-1\}$ and $d_\textit{max}$ is the disparity range\cite{hirschmuller2008evaluation}. Different kinds of cost functions can be used to compute the matching cost volume $C(\textbf{p},d)$.\\
\indent \textbf{Sum of absolute differences (SAD)} adds 
the absolute differences of intensities over all the pixels in a square window $\textit{N}_\text{p}$ centered at the pixel of interest \textbf{p}. For each pixel \textbf{q} in $\textit{N}_\text{p}$ in the base image, the absolute difference is computed by comparing the value of \textbf{q} with its corresponding pixel \textbf{q}-\textit{d} in the match image. The matching cost is computed as
\begin{equation}
    \textit{C}_\textit{SAD}(\textbf{p},\textit{d})=\textstyle\sum_{\textbf{q}\in \textit{N}_\text{p}}|I_\textit{b}(\textbf{q})-I_\textit{m}(\textbf{q}-d)|,
	\label{sad}
\end{equation}
where $I_\textit{b}(\textbf{q})$ and $I_\textit{m}(\textbf{q}-d)$ are the values of corresponding pixels in the base and match images, respectively.\\
\indent \textbf{Zero-mean sum of absolute differences (ZSAD)} subtracts the mean intensity of the window $\textit{N}_\text{p}$ from each intensity inside the window before computing the sum of absolute differences:
\begin{equation}
    \textit{C}_\textit{ZSAD}(\textbf{p},\textit{d})=\textstyle\sum_{\textbf{q}\in \textit{N}_\text{p}}|I_\textit{b}(\textbf{q})-\bar{I}_\textit{b}(\textbf{p})-(I_\textit{m}(\textbf{q}-d)-\bar{I}_\textit{m}(\textbf{p}-d))|,
	\label{zsad}
\end{equation}
where $\bar{I}_\textit{b}(\textbf{p})$ and $\bar{I}_\textit{m}(\textbf{p}-d)$ are the mean intensities of the windows centered at pixel \textbf{p} and \textbf{p}-\textit{d}, respectively.\\
\indent \textbf{Census transform} indicates the relative order of intensities in a local window centered at the pixel of interest, not the intensity values themselves\cite{zabih1994non}. It encodes each window into a bit string as follows:
\begin{equation}
    \textit{CT}(\textbf{p})=\otimes_{\textbf{q}\in \textit{N}_\text{p}}\Phi(I(\textbf{p}),I(\textbf{q})),
	\label{censustransform}
\end{equation}
where $\otimes$ denotes the bit-wise concatenation, and $I(\textbf{p})$ and $I(\textbf{q})$ are the values of the central pixel and the neighboring pixel in the window $\textit{N}_\text{p}$, respectively. $\Phi(i,j)$ is set to 1 if $i>j$, or 0 otherwise. After encoding, the matching cost is calculated through the Hamming distance between bit strings of corresponding pixels in the image pair, which is defined as the number of bits that are not equal, as shown in Eq. \ref{census}:
\begin{equation}
    \textit{C}_\textit{census}(\textbf{p},\textit{d})=\mathbb{H}(\textit{CT}_\textit{b}(\textbf{p}),\textit{CT}_\textit{m}(\textbf{p}-\textit{d})),
	\label{census}
\end{equation}
where $\textit{CT}_\textit{b}(\textbf{p})$ and $\textit{CT}_\textit{m}(\textbf{p}-\textit{d})$ are transformed bit strings of corresponding pixels in the image pair.\\
\indent \textbf{Rank transform} is also based on the relative ordering of local intensities \cite{zabih1994non}, defined as the number of pixels in the window $\textit{N}_\text{p}$ with an intensity less than the central pixel \textbf{p}:
\begin{equation}
    \textit{RT}(\textbf{p})=\|\{\textbf{q}\in \textit{N}_\text{p}|I(\textbf{q})<I(\textbf{p})\}\|.
	\label{ranktransform}
\end{equation}
The matching cost is then computed with the absolute difference between the transformed values of corresponding pixels in base and match images, as shown in Eq. \ref{rank}: 
\begin{equation}
    \textit{C}_\textit{rank}(\textbf{p},\textit{d})=|\textit{RT}_\textit{b}(\textbf{p})-\textit{RT}_\textit{m}(\textbf{p}-\textit{d})|.
	\label{rank}
\end{equation}
\subsection{Cost Aggregation}
To smooth the matching cost volume, SGM aggregates costs along independent paths, as shown in Fig. \ref{SGM flow}. For each path, SGM optimizes a minimization problem recursively based on the pathwise energy function, defined as
\begin{equation}
    \begin{aligned}
    L_r(\textbf{p},d) &=C(\textbf{p},d)+\min\left\{ \vphantom{y^2}
    \begin{aligned}
    &L_r(\textbf{p}-\textbf{r},d), \\
    &L_r(\textbf{p}-\textbf{r},d\pm1)+P_1,\\
    &\min_i L_r(\textbf{p}-\textbf{r},i)+P_2
    \end{aligned} 
    \right\} \\
    &\quad-\min_i L_r(\textbf{p}-\textbf{r},i),
    \end{aligned}
    \label{aggregation}
\end{equation}
where $L_r(\textbf{p},d)$ denotes the path cost of pixel \textbf{p} at disparity \textit{d} in direction \textbf{r} and $C(\textbf{p},d)$ represents the matching cost computed in Section \ref{cost functions}. The second term represents the recursive aggregation from the previous pixel $\textbf{p}-\textbf{r}$ in direction \textbf{r}. Penalties $P_1$ and $P_2$ are added for disparity adaption in slanted/discontinuous surfaces\cite{boykov2001fast}. The minimum path cost in the third term is subtracted to avoid a very large value. The aggregated cost is then computed by summing all the path costs, i.e., $S(\textbf{p},d)=\sum_r L_r(\textbf{p},d)$. 

\subsection{Disparity Computation and Refinement}\label{dispComp}
The disparity $D_b(\textbf{p})$ of each pixel \textbf{p} in the base image is computed using a winner-takes-all (WTA) strategy\cite{hirschmuller2007stereo}, which selects the disparity with the minimum aggregated cost, i.e., $D_b(\textbf{p})=\min_d S(\textbf{p},d)$. After computation, we provide several optional disparity refinement methods to remove outliers.\\
\indent \textbf{Left-right (L-R) consistency check} removes mismatching and occluded pixels\cite{hirschmuller2007stereo}. The disparity of the pixel \textbf{p} in the based image, i.e.,
$D_b(\textbf{p})$, is set to invalid if it differs from the disparity of the corresponding pixel $\textbf{p}'$ in the match image, i.e., $D_m(\textbf{p}')$, by more than 1 px. To obtain the disparity map of the match image, an efficient way is to reuse the aggregated cost volume for the base image based on $D_m(\textbf{p}')=\text{arg}\min_d S(\textbf{p}'+d,d)$. A more accurate method is to perform cost computation and aggregation from scratch and switch the position of the base and match images. Both methods are supported in our library.\\
\indent Finally, we also provide a \textbf{median filter} with a user-defined window size to remove outliers and smooth the disparity map.
\section{Optimization and Automation}\label{optimization techniques}
In this section, we will present our optimization and automation techniques which are applied in FP-Stereo to boost performance, save resources and improve usability. \\ 
\begin{figure}
  \centering
  \includegraphics[width=0.98\columnwidth]{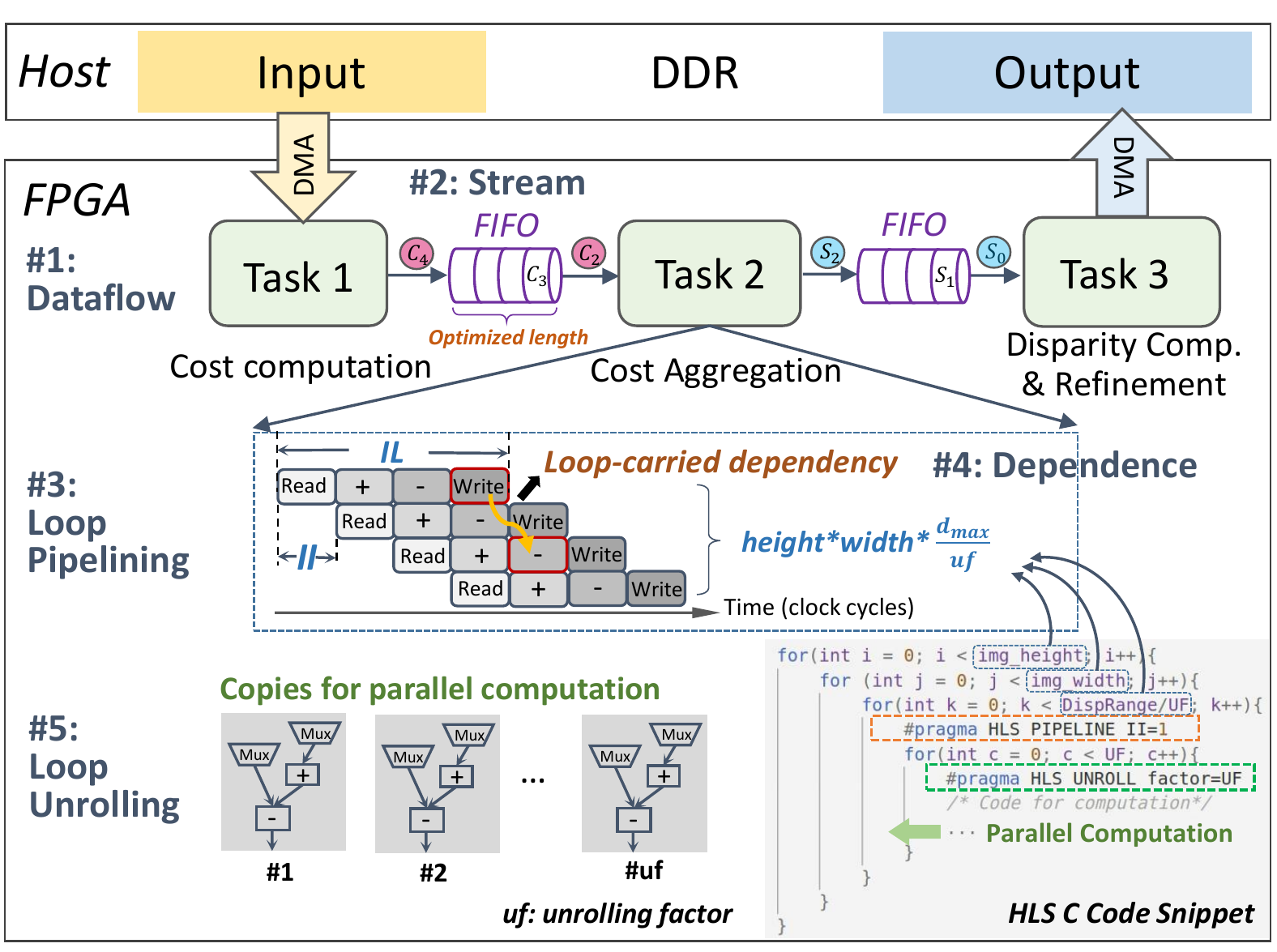}
  \caption{Optimization with HLS pragmas.}
  \label{optimization}
  \vspace{-0.1cm}
\end{figure}
\begin{figure}
  \centering
  \includegraphics[width=\columnwidth]{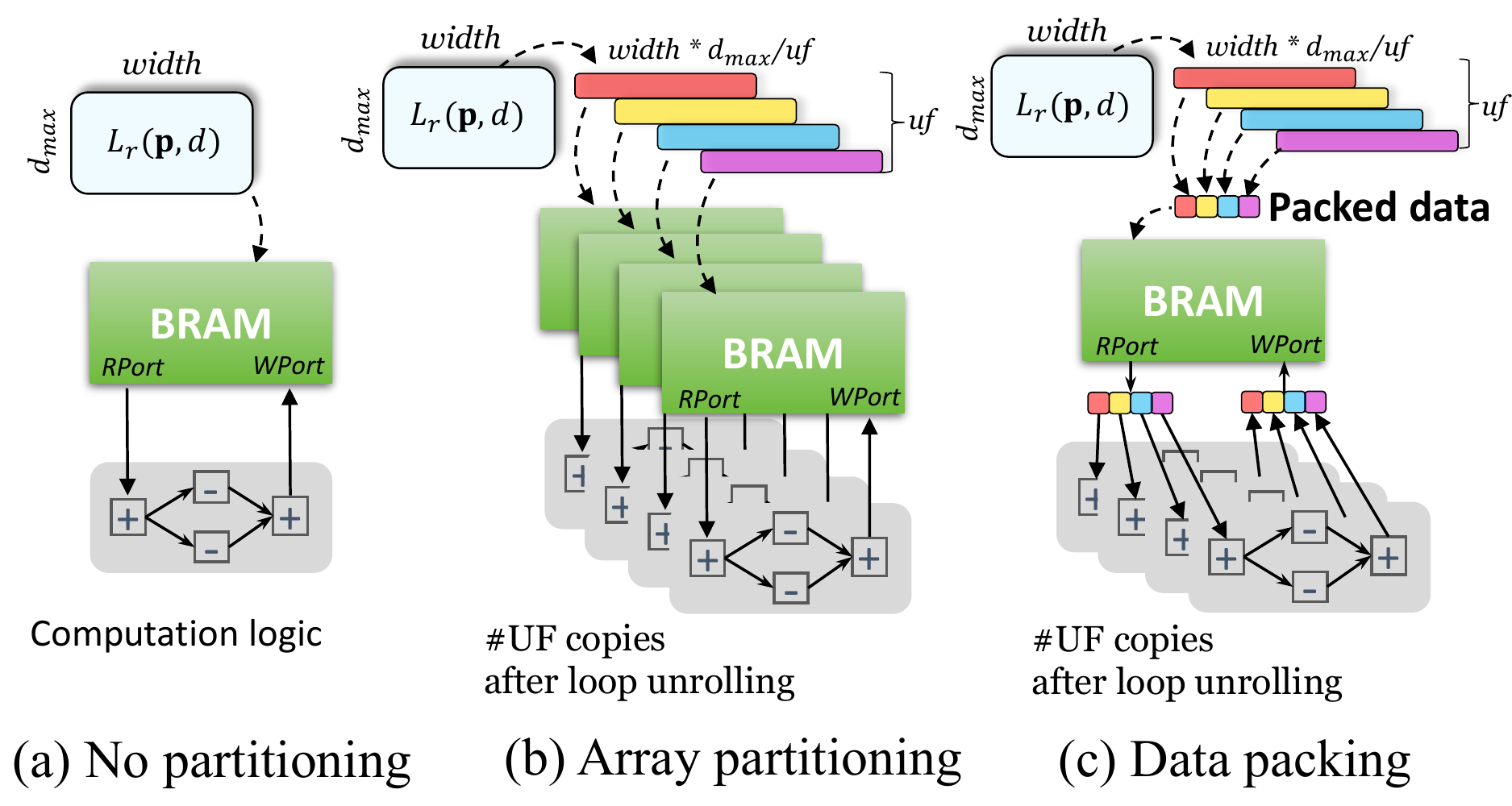}
  \caption{Memory optimization.}
  \label{mem}
\end{figure}
\begin{figure}
  \centering
  \includegraphics[width=0.99\columnwidth]{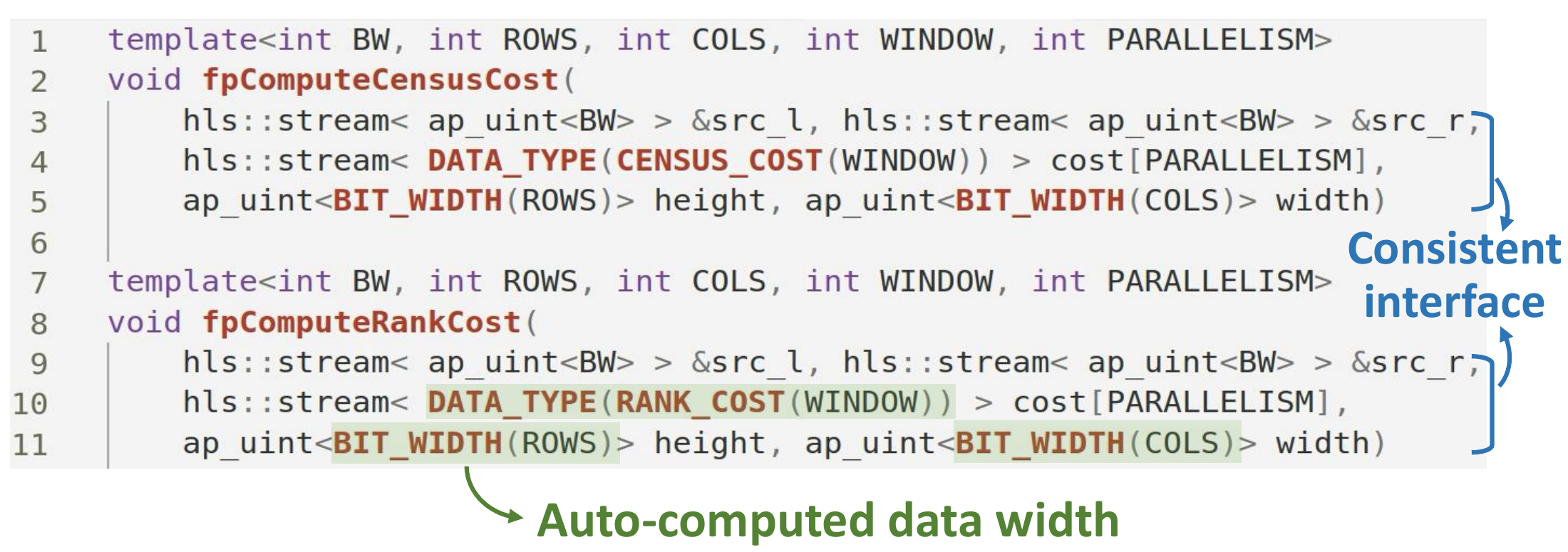}
  \caption{Code snippet of the library.}
  \label{interface}
\end{figure} 
\indent \textbf{Well-set HLS pragmas.} To boost the performance and exploit the parallelism, we optimize the functions in the library with various HLS pragmas, as shown in Fig. \ref{optimization}. \textit{Dataflow} enables coarse-grained parallelism among functions and loops. By inserting FIFOs between each module, data is sent to the consumer immediately once processed by the producer, and different modules execute concurrently. In this case, the overall latency equals the latency of the slowest module \cite{zhao2019performance}. 
To keep a high data rate while considering limited on-chip memory resources, FIFO lengths are set as short as possible using the \textit{stream} pragma, provided that the data path is not stalled. \textit{Loop pipelining} and \textit{loop unrolling} exploit fine-grained parallelism within loops. A pipelined loop processes new inputs at an initiation interval ($\textit{II}$), constrained by the loop-carried dependency\cite{zhao2017comba,zhao2019performance}. 
To eliminate harmful data dependencies, the \textit{dependence} pragma is applied to avoid the false dependence analysis of HLS tools which may happen under complicated scenarios. Moreover, the hardware implementation should also be well designed to resolve harmful dependence, which will be discussed in Section \ref{implementation}. \textit{Loop unrolling} allows concurrent execution of independent iterations by creating copies of the loop body, 
and is inserted in the inner-most loop considering both efficiency and hardware cost. In our case, the latency of each module (clock cycles) can be estimated using
\begin{equation}
    \textit{Cycle}=\textit{IL}+\textit{II}\times (\textit{height}\times \textit{width}\times \frac{\textit{d}_\textit{max}}{\textit{uf}}-1),
	\label{pipeline_cycle}
\end{equation}
where \textit{IL} is the iteration latency, \textit{height} is the image height, \textit{width} is the image width, $\textit{d}_\textit{max}$ is the disparity range and \textit{uf} is the unrolling factor in the disparity dimension.\\
\indent \textbf{Memory optimization.} Simultaneous memory accesses are critical for parallel computation, but may not be ensured by the on-chip memories on FPGA (i.e., BRAMs) due to the limited number of read and write ports. Figure \ref{mem} presents how we utilize BRAMs efficiently to maximize performance. We take a row of data with the size $\textit{width}*\textit{d}_\textit{max}$ from the path cost volume $L_r(\textbf{p},d)$ as an example. The data is stored in an array which is mapped to BRAMs. When the loop is unrolled, to enable concurrent reads and writes required by the copies of computation logic, the \textit{array partitioning} pragma divides the array into multiple partitions and stores each partition in different BRAMs, boosting the parallelism while incurring multi-fold memory usage. To save memory resources, we pack the elements required by the parallel computation and store the packed data in a single memory instance, as shown in Fig. \ref{mem}(c). The required data can still be accessed quickly by fetching the packed data and feeding different bits to the corresponding computation logic. Through \textit{data packing}, the on-chip memories on FPGA can be fully utilized and the memory usage is reduced by 50\% compared to \textit{array partitioning}. Note that we still apply \textit{array partitioning} to small arrays which will be mapped to registers.\\
\indent \textbf{Auto-computed data width and consistent interface.} In general C code, the native data types are all on multiple bytes, which can result in inefficient hardware. In HLS, users can specify an arbitrary data width for each variable. To improve usability, FP-Stereo automatically specifies the optimal data width for each variable without influencing the accuracy. Based on closed-form expressions, the data widths are computed automatically during compilation. Specifically, the data widths depend on the algorithmic characteristics and parameters, such as the type of cost functions and the window size, as shown in Fig. \ref{interface}. Our flexible optimized data types significantly reduce resource usage, resulting in a faster circuit and allowing for a higher clock frequency. Moreover, function templates in the same stage of the pipeline are implemented with the consistent interface in a unified format, ensuring the scalability of our library.\\
\indent \textbf{Automatic Code Generation.} To quicken the FPGA development cycle, FP-Stereo can automatically generate the complete stereo matching pipeline based on our optimized HLS templates. Given user-specified algorithmic parameters, FP-Stereo generates function calls, sets parameters and inserts buffers 
between functions to form a deep pipeline. The input and output images are streamed into and out of FPGA through the direct memory access (DMA) engine. Besides the accelerator code, FP-Stereo also generates the host code to allocate/free the DDR memory space and invoke the accelerator.
\section{Implementation Details}\label{implementation}
\subsection{Matching Cost Computation}
\begin{figure}
  \centering
  \includegraphics[width=\columnwidth]{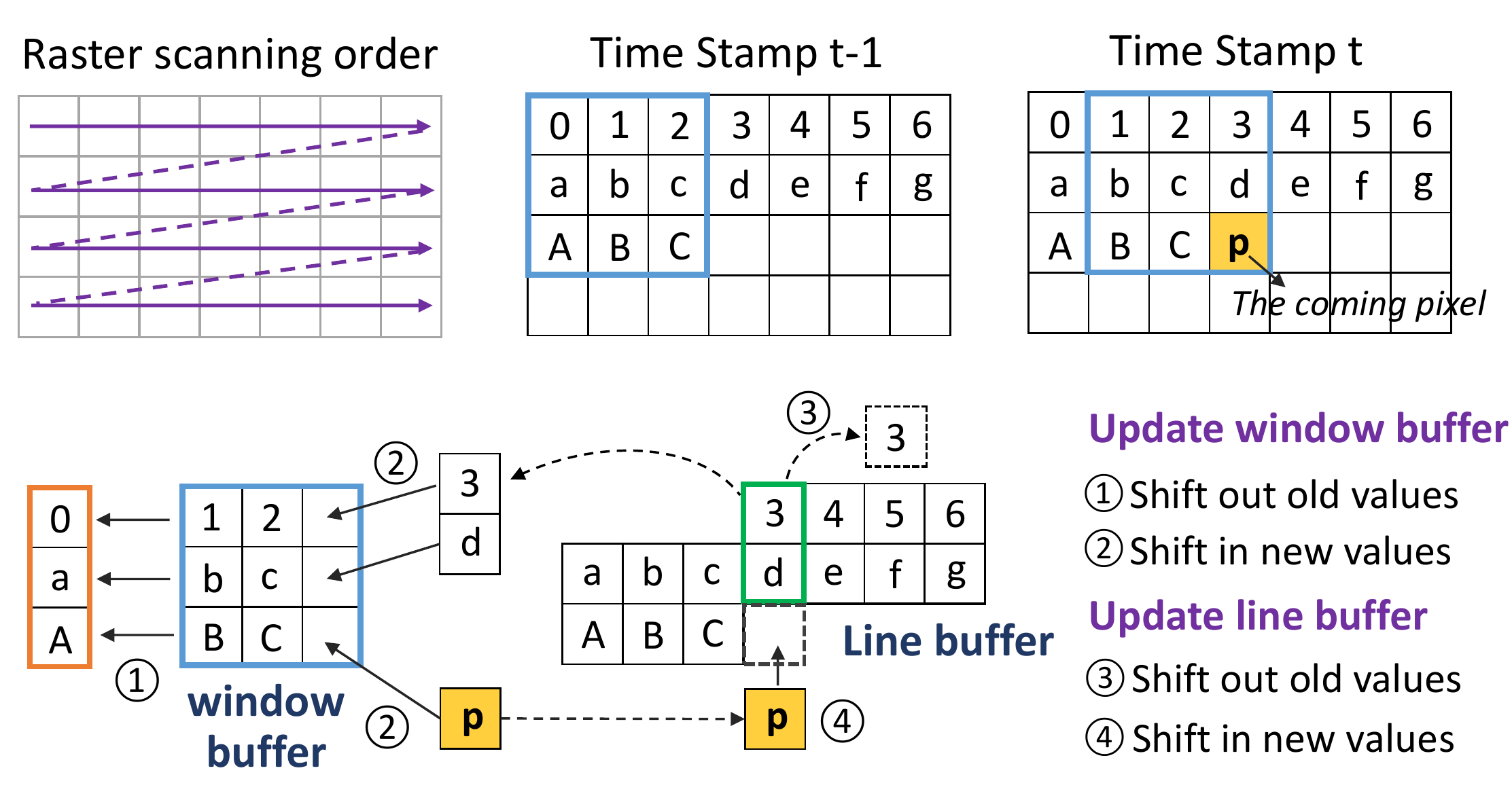}
  \caption{Buffer reuse strategy. A $3\times 3$ window buffer and a $2\times 7$ line buffer are presented for illustration.}
  \label{window_line}
\end{figure}
The inputs of the matching cost computation stage are the 8-bit intensities of base and match images and the output is the matching cost volume. As discussed in Section \ref{cost functions}, cost functions rely on window-based computation. Neighboring pixels in a local window are collected to compute the matching costs of the central pixel. As shown in Fig. \ref{window_line}, input image intensities are transferred to FPGA every clock cycle in the raster scanning order and the windows to be processed at time stamps \textit{t}-1 and \textit{t} are highlighted. The notations represent the pixels that have been streamed in FPGA and the pixel \textbf{p} is the coming pixel at clock cycle \textit{t}. The image data is transferred to FPGA continuously in a single burst after the host receives the transfer request. Each piece of data is transferred once and needs to be stored in the local memory on FPGA for reuse.\\
\indent We utilize \textit{window and line buffers} for data reuse, as shown in Fig. \ref{window_line}. A window buffer stores neighboring values in a local window and a line buffer stores multiple rows of values. At each clock cycle, the window buffer is updated by shifting out the left-most column of data and shifting in the data part from the line buffer and part from the coming pixel. The line buffer is then updated to store the most recently visited rows of data by shifting out the old value and shifting in the coming pixel in the same column. This date reuse strategy efficiently utilizes on-chip memories by storing the necessary data required by the computation in the raster order rather than the whole image.\\ 
\begin{figure}
  \centering
  \includegraphics[width=\columnwidth]{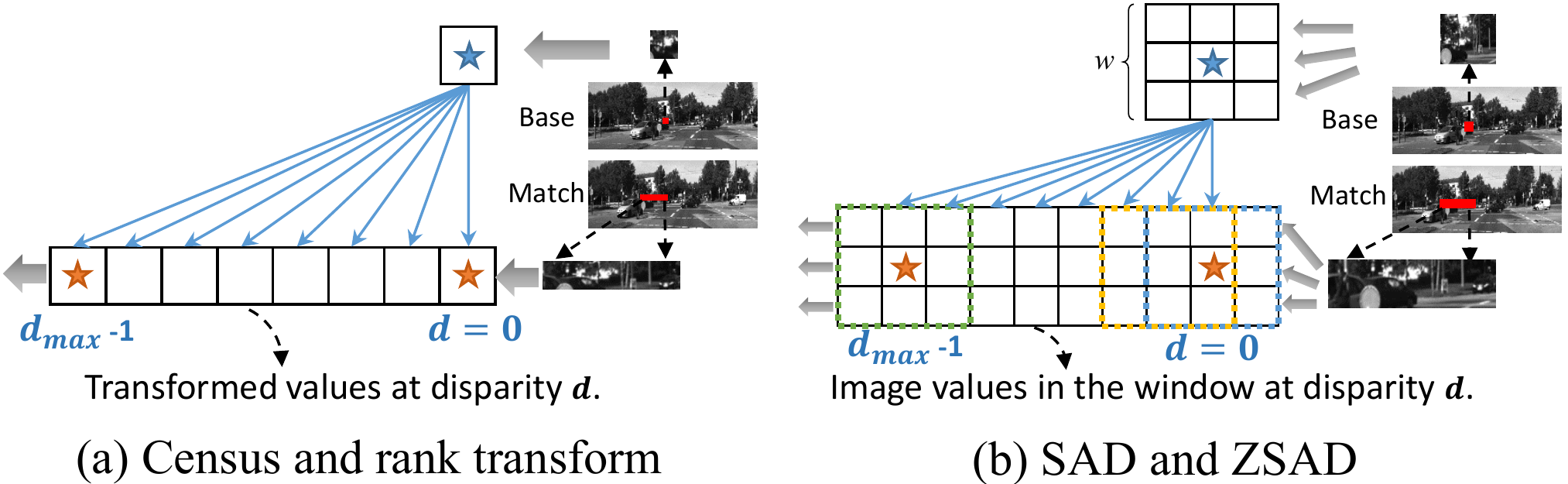}
  \caption{Matching cost computation for the base image}
  \label{cost_computation}
\end{figure}
\begin{figure}
  \centering
  \includegraphics[width=\columnwidth]{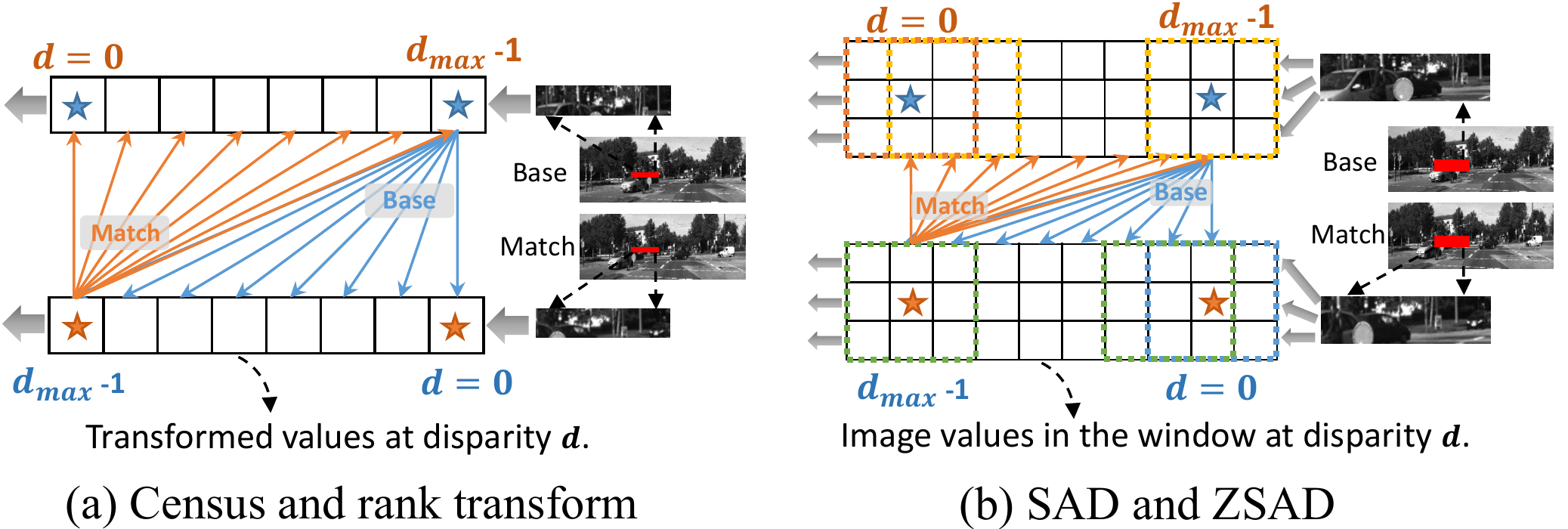}
  \caption{Matching cost computation for base \& match images}
  \label{cost_computation_lr}
\end{figure}
\indent Based on window and line buffers, the neighboring pixels of the target pixel within a local window can be accessed immediately for the matching cost computation in the raster scanning order. For census and rank transform, the neighboring pixels are compared to the central pixel in parallel. After transformation, as shown in Fig. \ref{cost_computation}(a), the hamming distance or the absolute difference is computed between the transformed values of the target pixel in the base image and its corresponding pixels at each disparity in the match image. Then the FIFO is updated by shifting left to remove the unused data and store the new incoming data to get prepared for the next computation. For SAD and ZSAD, the window centered at the target pixel is directly compared to the windows centered at corresponding pixels at each disparity. To save resources, we use a window buffer with the size $\textit{w}*(d_\textit{max}+\textit{w}-1)$ to store the windows of the match image, which is updated as shown in Fig. \ref{cost_computation}(b). In the case of the L-R consistency check, the matching costs of the match image are computed simultaneously, as is shown in Fig. \ref{cost_computation_lr}. Note that the computation for the match image is slower than that of the base image for $d_\textit{max}-1$ pixels. FP-Stereo supports both types of matching cost computation in Fig. \ref{cost_computation} and Fig. \ref{cost_computation_lr} for each cost function. The matching cost computation stage executes as an efficient pipeline with $\textit{II}=1$, sending the matching costs to the next stage every clock cycle.
\subsection{Cost Aggregation}
In this stage, the matching costs are accumulated along several paths and the path costs are summed to calculate the aggregated cost.
\begin{figure}
  \centering
  \includegraphics[width=\columnwidth]{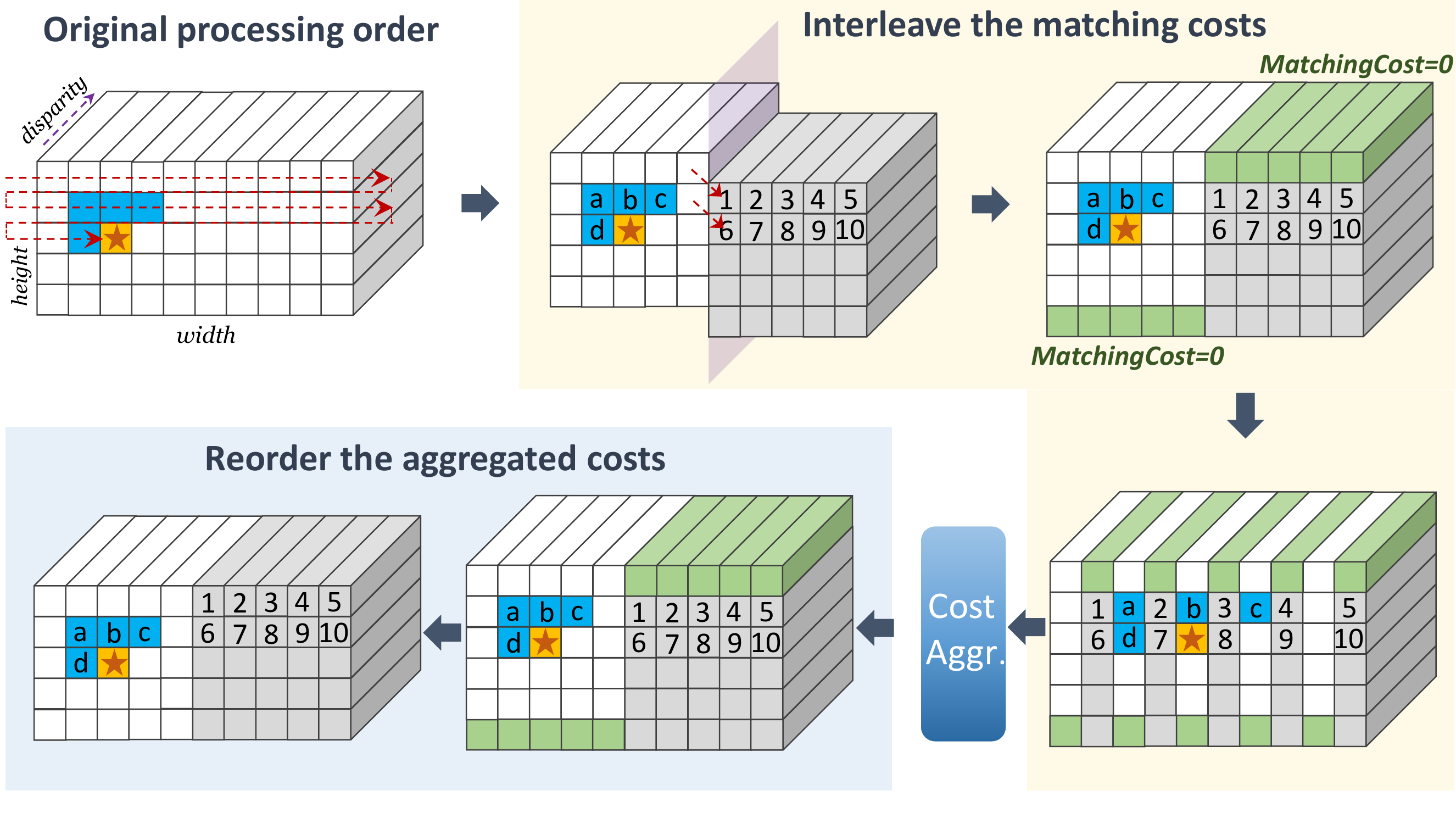}
  \caption{Interleaving and reordering for cost aggregation. The target pixel and its preceding pixels are highlighted in orange and blue, respectively. }
  \label{cost_aggregation}
\end{figure}
The independent pathwise accumulation enables concurrent computation, which may not be fully exploited due to limited on-chip memory resources. To solve this issue, parallelism can be achieved by accumulating along the paths which are consistent with the scanning order. As shown in Fig. \ref{cost_aggregation}, since pixels are processed in the raster order, the path costs of the preceding pixels (blue) in the directions of $0^\text{o}$, $45^\text{o}$, $90^\text{o}$ and $135^\text{o}$ are available when performing the accumulation at the target pixel (orange).
For the $45^\text{o}$, $90^\text{o}$ and $135^\text{o}$ directions, the accumulation at the target pixel relies on the path costs computed in the previous row.
Line buffers are used to store the path costs for each direction and updated with the most recently visited row. To save memory resources, the path costs which are computed in parallel are packed. For the direction of $0^\text{o}$ which exactly follows the raster order, the path costs are stored in registers which are updated at each pixel. When traversing each pixel, the computations in the four paths are performed concurrently and the path costs are summed immediately after computation.\\
\indent The cost aggregation stage outputs aggregated costs of each pixel at different disparities at an initiation interval (\textit{II}). In the direction of $0^\text{o}$, each pixel has to wait for the computation of the left pixel to finish, leading to a tight dependence and increasing $\textit{II}$. For other directions, the computations of preceding pixels are completed earlier by one row, which does not influence the \textit{II}. To minimize $\textit{II}$, we propose a novel \textit{interleaving and reordering} strategy to enlarge the distance between dependent pixels, as illustrated in Fig. \ref{cost_aggregation}. The matching cost volume is divided in half, and the latter half is shifted down to be interleaved with the former half. The missing locations are padded with zeros. In this way, the matching costs in the latter half are still processed after the costs of the former half in the same row, maintaining their relative positions in the raster scanning order rather than affecting the original algorithm behavior.
Following the rescheduled order, the pixels with tight dependence are interleaved and the computation at grey positions can be performed concurrently while the white positions wait for the results of preceding pixels. The aggregated costs are then reordered to follow the original order and transferred to the next stage with $\textit{II}=1$.\\
\indent Four FIFOs are used for interleaving and reordering, and \textit{data packing} is used to save memory resources. Additional registers are used to store results computed at the cutting edge. The number of paths to aggregate matching costs influences both \textit{accuracy} and \textit{speed}. 
The four-path aggregation achieves high hardware efficiency, while the accuracy may be affected due to the missing backward information from other directions (e.g., $180^\text{o}$). Based on our scalable library-based framework, we will support the cost aggregation in more directions in the near future.  
\subsection{Disparity Computation and Refinement}
\begin{figure}
  \centering
  \includegraphics[width=\columnwidth]{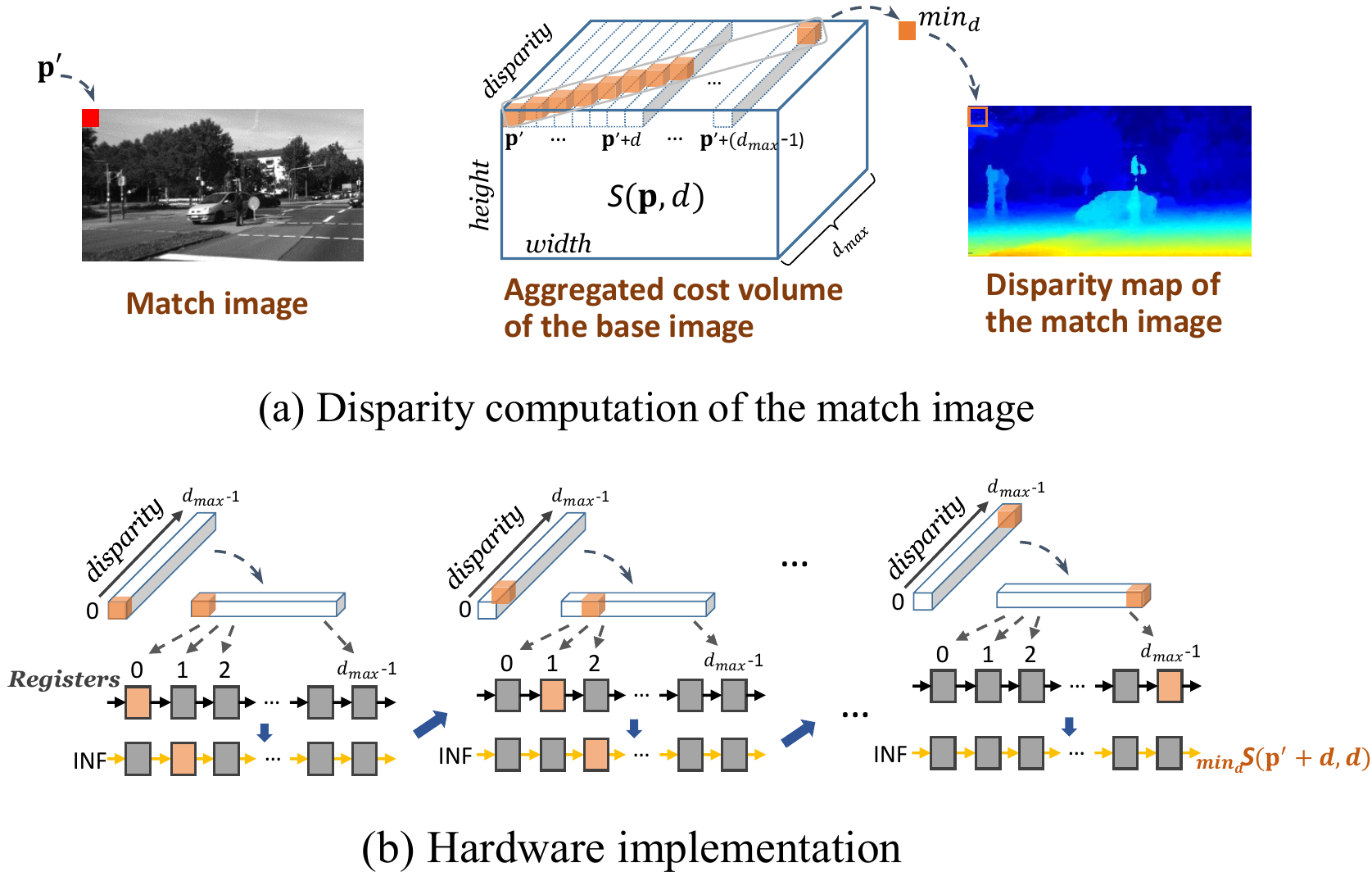}
  \caption{Disparity computation for the match image}
  \label{consistency}
\end{figure}
The disparity of each pixel is computed using a multi-level multiplexer which selects the disparity with the minimum aggregated cost. The median filter is implemented based on window and line buffers. For the L-R consistency check, the disparity map of the match image, $D_m$, is required to compare with that of the base image. FP-Stereo provides two methods to compute $D_m$, as discussed in Section \ref{dispComp}. 
Figure \ref{consistency} illustrates the method of reusing the aggregated cost volume of the base image and our hardware implementation. Given a pixel $\textbf{p}'=(x,y)$ in the match image, its disparity is computed by comparing the aggregated costs of pixels $\textbf{p}',...,\textbf{p}'+(d_\textit{max}-1)$ in the base image, at disparities $0,...,d_\textit{max}-1$, respectively. 
As shown in Fig. \ref{consistency}(b), when the aggregated costs of each pixel come in the raster order, the registers are updated with the lower values after comparison to the costs at each disparity, and then shifted right for comparison with the aggregated costs of the next pixel. For illustration, the aggregated costs involved in the disparity computation for $\textbf{p}'$ in Fig. \ref{consistency}(a) are also highlighted in orange in Fig. \ref{consistency}(b). After $d_\textit{max}$ comparisons, the disparity of $\textbf{p}'$ is computed by recording the corresponding disparity of the minimum aggregated cost stored in the rightmost register. Note that the computations for other pixels are performed concurrently and the results are updated in the grey registers. Therefore, the disparities can be produced every clock cycle. For the method of computing from scratch, the cost aggregation function is instantiated twice, for the base and match images, respectively. Then the base and match disparity maps can be computed using their individual aggregated cost volume. This 
method increases the accuracy further but doubling the resource usage.
The appropriate approach can be chosen depending on the practical design requirements posed by the trade-off between accuracy and hardware budgets.

\section{Experimental Results}
To investigate the trade-off among \textit{accuracy}, \textit{speed} and \textit{resource usage}, we test different configurations of our library on FPGA, analyze the impact of various design choices and provide users guidance for the right design choice. 
We also compare our implementations with state-of-the-art FPGA and GPU solutions, demonstrating the superior performance of FP-Stereo in real-time embedded systems with strict \textit{power} constraints. The experiments are performed on Xilinx Ultrascale+ ZCU102 FPGA and the KITTI 2015 dataset \cite{menze2015object} is used for evaluation. The FPGA development toolkit we use is Xilinx SDSoC 2018.3. The power is measured on board using Xilinx Zynq UltraScale+ MPSoC Power Advantage tool\cite{xilinxpowertool}.
\subsection{Analysis on the impact of design choices}
We evaluate 576 configurations each of which includes a choice of the cost function and its window size ($5\times5$, $7\times7$), the disparity range (64, 128), the unrolling factor in the disparity dimension (4, 8, 16, 32), image resolution (1242*374, 900*260) and refinement methods.
\textit{Accuracy} is measured by the percentage of erroneous pixels on non-occluded areas (D1-all error from KITTI), \textit{speed} is reported through the execution time (s) of processing one image pair, and \textit{resource} \textit{usage} is evaluated by averaging the utilization ratios of each FPGA component, i.e., BRAMs, DSPs, LUTs and flip-flops (FFs). For the analysis in this section, the results of execution times and resources are from HLS reports.
\begin{figure}
  \centering
  \includegraphics[width=0.98\columnwidth]{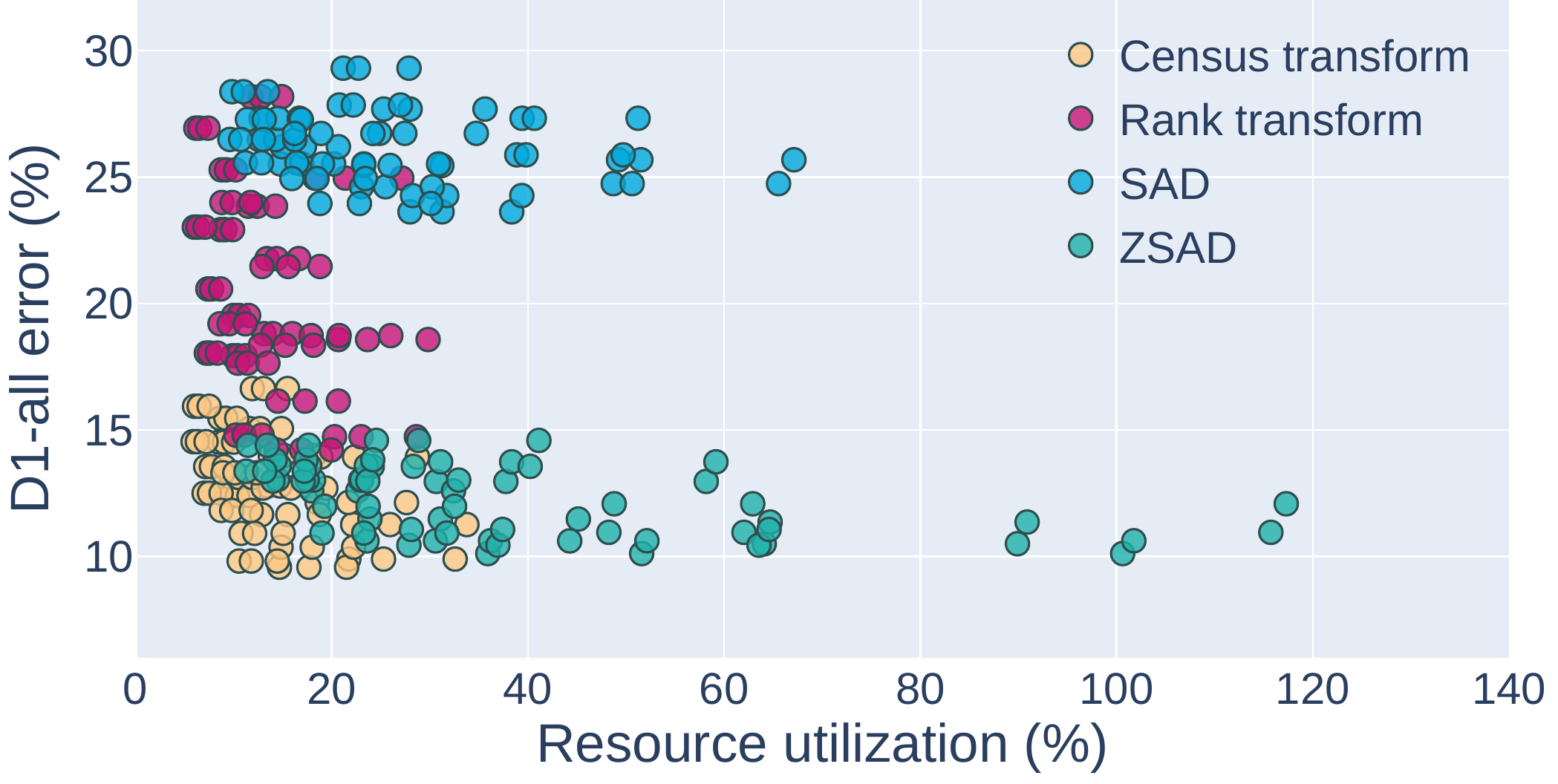}
  \caption{D1-all error vs resource utilization colored by different cost functions.}
  \label{cost_hls}
\end{figure}
\begin{figure}
  \centering
  \includegraphics[width=0.98\columnwidth]{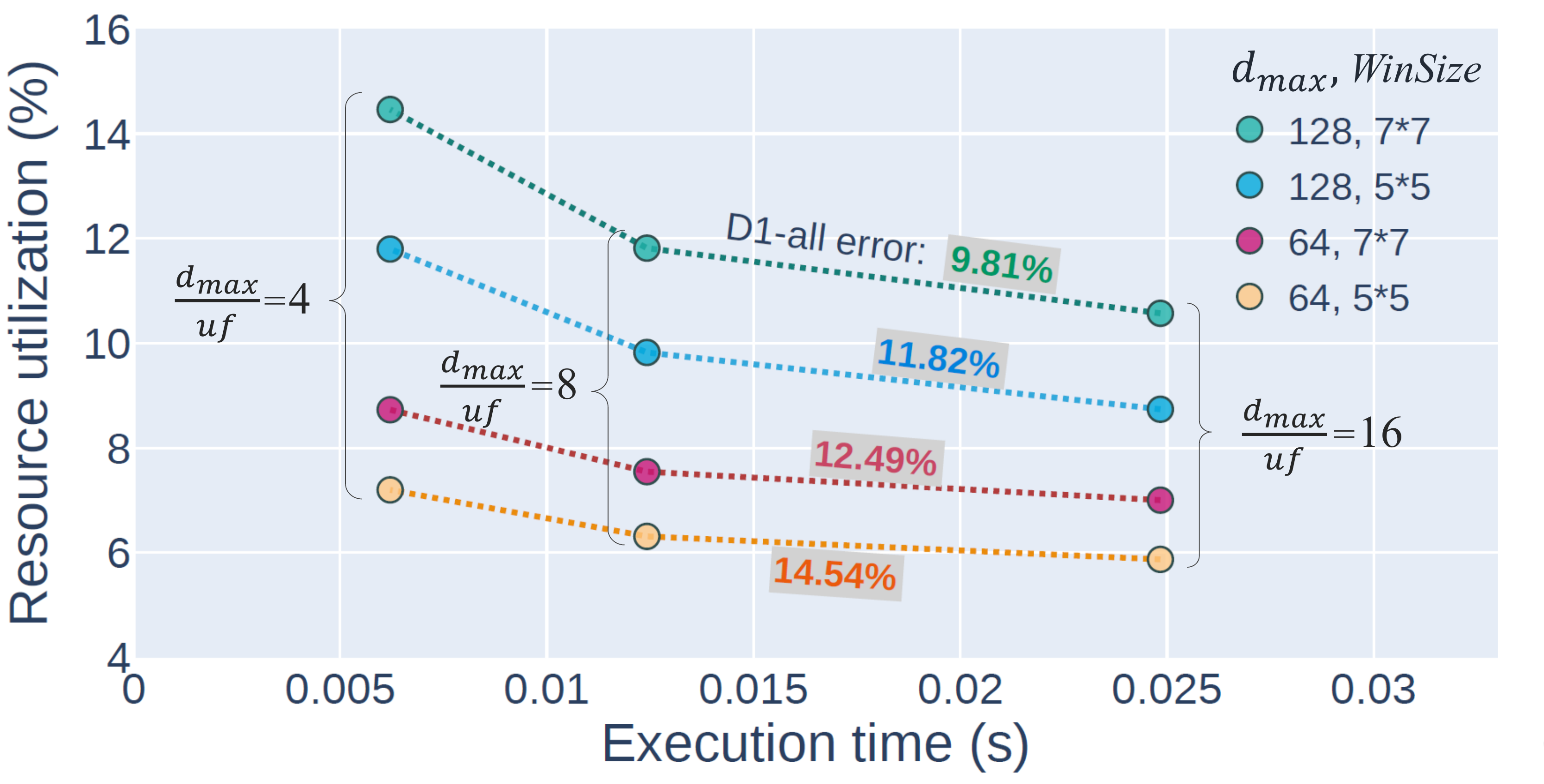}
  \caption{Comparison given different $\textit{WinSize}$, $d_\textit{max}$ and $\textit{uf}$. Configuration setting: census transform, image resolution of 1242*374, median filter and no L-R consistency check.}
  \label{dmax_hls}
\end{figure}
\\
\textbf{The impact of adjusting the cost function.} The execution time is computed through dividing the latency (clock cycles) by the frequency (300\text{MHz}). The latency can be estimated with Eq. \ref{pipeline_cycle} which can be used for theoretical analysis. When adjusting the type of cost function, the execution times are close with negligible differences caused by iteration latency \textit{IL}, which depends on the latencies of computing operations in each formulation in Section \ref{cost functions}. Figure \ref{cost_hls} plots all the tested configurations based on \textit{accuracy} vs \textit{resource utilization}, categorized by different cost functions. Census transform and ZSAD achieve higher accuracy and SAD performs the worst. Rank transform exhibits a large variance in \textit{accuracy} and some configurations lead to small errors. Compared to SAD and ZSAD, census and rank transform consume fewer resources and are more suitable to be mapped onto hardware. This is because census and rank transforms require smaller bit widths to represent matching costs without losing information, leading to less consumption of computation logic and memories. Compared to other methods, census transform is the most robust and hardware-efficient method considering both \textit{accuracy} and \textit{resource utilization}.
\begin{table*}
\begin{center}
\begin{tabular}{|c|c|c|c|c|c|c|c|c|c|c|c|c|}
\hline
    \multirow{2}*{\footnotesize{Method}} & \multirow{2}*{\footnotesize{Config.}} & \multicolumn{3}{c|}{\footnotesize{Accuracy}} & \multicolumn{2}{c|}{\footnotesize{Speed}} &
    \multicolumn{4}{c|}{\footnotesize{Resource Utilization}}&
    \multirow{2}*{\makecell{\footnotesize{Power }\footnotesize{(W)}}}
    & 
   \multirow{2}*{\makecell{\footnotesize{Energy }\footnotesize{(J)}}}\\
\cline{3-11}
      & &\footnotesize{D1-bg} & \footnotesize{D1-fg} & \footnotesize{D1-all} & \footnotesize{Runtime (s)} & \footnotesize{FPS} & \footnotesize{BRAM} & \footnotesize{DSP} & \footnotesize{LUT} & \footnotesize{FF} & 
      & \\
\hline\hline
\multirow{2}*{\footnotesize{xfOpenCV\cite{xilinxxfopencv}}} & \footnotesize{S0} & \footnotesize{15.51\%} & \footnotesize{27.42\%} & \footnotesize{17.31\%} &
\footnotesize{0.0124} & \footnotesize{81} & \footnotesize{13.8\%} & \footnotesize{3.0\%} & \footnotesize{16.7\%} & \footnotesize{5.7\%} & \footnotesize{{3.6}} & \footnotesize{8.9} \\ 
& \footnotesize{S1} & \footnotesize{13.77\%} & \footnotesize{21.85\%} & \footnotesize{14.99\%} &
\footnotesize{0.0124} & \footnotesize{81} & \footnotesize{24.3\%} & \footnotesize{5.6\%} & \footnotesize{25.2\%} & \footnotesize{7.3\%} & - & -\\ 
\hline
\multirow{5}*{\footnotesize{FP-Stereo}} & 
\footnotesize{C0} & \footnotesize{12.62\%} & \footnotesize{{25.31\%}} & \footnotesize{14.54\%} & \footnotesize{\textbf{0.0062}} & \footnotesize{\textbf{161}} & \footnotesize{\textbf{9.2\%}} & \footnotesize{\textbf{2.9\%}} & \footnotesize{\textbf{13.4\%}} & \footnotesize{\textbf{3.2\%}} & \footnotesize{\textbf{4.1}} & \footnotesize{\textbf{5.1}}\\ 
& \footnotesize{C1} & \footnotesize{10.53\%} & \footnotesize{\textbf{19.04\%}} & \footnotesize{11.82\%} & \footnotesize{\textbf{0.0062}} & \footnotesize{\textbf{161}} & \footnotesize{{16.9\%}} & \footnotesize{{5.5\%}} & \footnotesize{{20.1\%}} & \footnotesize{{4.7\%}} & \footnotesize{5.7} & \footnotesize{7.1}\\ 
& \footnotesize{C2} & \footnotesize{8.16\%} & \footnotesize{19.11\%} & \footnotesize{9.81\%} & \footnotesize{\textbf{0.0062}} & \footnotesize{\textbf{161}} & \footnotesize{19.6\%} & \footnotesize{{5.5\%}} & \footnotesize{26.6\%} & \footnotesize{6.1\%} & \footnotesize{6.6} & \footnotesize{8.2}\\ 
&\footnotesize{C3} & \footnotesize{7.78\%} & \footnotesize{19.61\%} & \footnotesize{9.57\%} & \footnotesize{0.0068} & \footnotesize{147} & \footnotesize{19.9\%} & \footnotesize{5.6\%} & \footnotesize{52.8\%} & \footnotesize{7.7\%} &\footnotesize{6.7} & \footnotesize{9.1} \\ 
&\footnotesize{C4} & \footnotesize{\textbf{7.69\%}} & \footnotesize{20.37\%} & \footnotesize{\textbf{9.49}\%} & \footnotesize{0.0068} & \footnotesize{147} & \footnotesize{38.7\%} & \footnotesize{10.8\%} & \footnotesize{68.7\%} & \footnotesize{12.1\%} & \footnotesize{9.8}&\footnotesize{13.3}\\ 
\hline
\end{tabular}
\end{center}
\caption{Comparison on FPGA. S0, C0: $\textit{WinSize}=5\times5$, $d_\textit{max}=64$, $\textit{uf}=16$, NLR. S1, C1: $\textit{WinSize}=5\times5$, $d_\textit{max}=128$, $\textit{uf}=32$, NLR.
C2-C4: $\textit{WinSize}=7\times7$, $d_\textit{max}=128$, $\textit{uf}=32$ and NLR, LR1 and LR2 are applied respectively. Median filter is applied to C0-C4.}
\label{FPGA comparisions}
\end{table*}
\\
\noindent\textbf{The impact of adjusting the window size ($\textit{WinSize}$), the disparity range ($d_\textit{max}$) and the unrolling factor ($\textit{uf}$).} We vary the three variables and plot the configurations in Fig. \ref{dmax_hls}, while fixing other design choices as in the configuration setting. 
The execution time is related to ${d_\textit{max}}/\textit{uf}$, as shown in Eq. \ref{pipeline_cycle}, which reflects the parallelism degree on hardware. With smaller ${d_\textit{max}}/\textit{uf}$, the parallelism is improved, leading to faster speeds while consuming more resources. The accuracy is positively correlated with $d_\textit{max}$ and $\textit{WinSize}$. As shown in Fig. \ref{dmax_hls}, with larger $d_\textit{max}$ and $\textit{WinSize}$, more image information is involved to reduce the estimation error, while incurring a heavier computation burden. This leads to more resource consumption to achieve the same speed.
\begin{figure}
  \centering
  \vspace{-0.2cm}
  \includegraphics[width=\columnwidth]{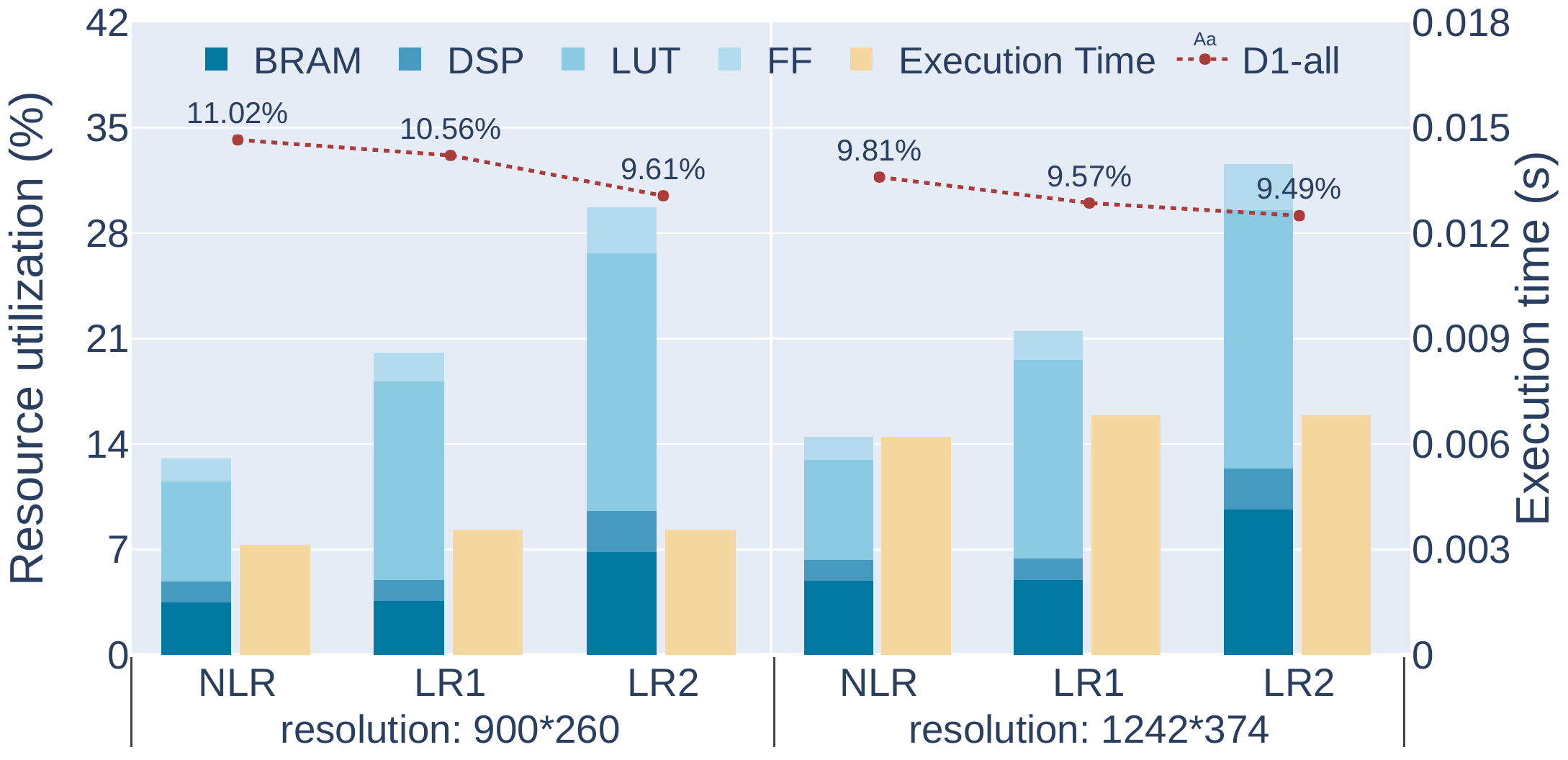}
  \caption{Comparison for L-R check methods and image size. Configuration setting: census transform, $\textit{WinSize}=7*7$, $d_\textit{max}=128$, $uf=32$ and median filter.}
  \label{lrcheck_hls}
\end{figure}
\\
\textbf{The impact of adjusting L-R check methods and image size.} In Fig. \ref{lrcheck_hls}, we compare various L-R check methods in terms of \textit{accuracy}, \textit{speed} and \textit{resource usage}, given different image resolutions. NLR represents that the L-R consistency check is not applied. For the L-R consistency check, LR1 and LR2 denote different methods of computing the match image disparities, i.e., computing using the aggregated costs of the base image and computing from scratch, respectively. Given the same resolution, both LR1 and LR2 improve the accuracy but consume more resources and slightly increase the runtime due to additional computation. Specifically, LR2 consumes more resources than LR1 because of the duplication of functions to exploit parallelism. Different image resolutions mainly impact the \textit{speed} and the BRAM usage, because the latency can be smaller given fewer pixels to be processed, and the length of line buffers is reduced given images with shorter image widths. 
\subsection{Comparison with the state-of-the-art on KITTI}
To illustrate the performance of FP-Stereo, we compare our configurations with state-of-the-art implementations. KITTI images with the full resolution of 1242*374 are tested. 
We evaluate our configurations on board and Table \ref{FPGA comparisions} lists the results tested on the Xilinx Ultrascale+ ZCU102 FPGA board. The \textit{accuracy} is measured on 200 KITTI image pairs following the default error criterion of KITTI. The \textit{speed} is evaluated with the runtime (s) of processing one image pair and frames per second (FPS). The \textit{power consumption} is measured using the Power Advantage Tool\cite{xilinxpowertool} and the \textit{energy} denotes the total energy consumption for the KITTI dataset with 200 image pairs, which is computed as $\textit{power}*\textit{runtime}*200$. Compared to the SGM configurations provided by Xilinx xfOpenCV\cite{xilinxxfopencv} (S0, S1) and given the same algorithmic parameters, FP-Stereo configurations (C0, C1) achieves 3.24\% lower error, 2x faster speed, 30\% less resource usage and 40\% less energy consumption. FP-Stereo can also provide more options, such as C2-C4, which can further improve the accuracy (6.08\% lower error) while maintaining a fast speed (1.8x) at the cost of more resources and energy. The different configurations provided by FP-Stereo demonstrate the flexibility of our library to meet different practical requirements and the capability of FP-Stereo to balance different design metrics for real-world embedded applications.\\
\begin{table}
\begin{center}
\begin{tabular}{|c|c|c|c|c|c|}
\hline
    \scriptsize{Method} & {\makecell{\scriptsize{Speed}\\(\scriptsize{FPS})}} & {\makecell{\scriptsize{Power}\\(\scriptsize{W})}} &
    {\makecell{\scriptsize{Energy}\\(\scriptsize{J})}} & {$\frac{\text{FPS}}{\text{Watt}}\ $}& {\scriptsize{Platform}}\\
\cline{2-6} 
\hline\hline
\scriptsize{OpenCV}\cite{opencv} & \scriptsize{0.27} & \scriptsize{1.6} & \scriptsize{1185} & 1x & \scriptsize{ARM Cortex-A53 CPU}\\ 
\scriptsize{SGM}\cite{chacon2013n} & \scriptsize{\textbf{238}} & \scriptsize{{101}} & \scriptsize{84.9} & \scriptsize{14x} & \scriptsize{Nvidia Titan X GPU}\\ 
\scriptsize{SGM}\cite{chacon2013n}& \scriptsize{29} & \scriptsize{11.7} & \scriptsize{80.7} & \scriptsize{15x} & \scriptsize{Nvidia JetsonTX2 GPU}\\ 
\scriptsize{FP-Stereo-C2} & \scriptsize{161} & \scriptsize{\textbf{6.6}} & \scriptsize{\textbf{8.2}} & \scriptsize{\textbf{145x}} & \scriptsize{Xilinx ZCU102 FPGA}\\ 
\hline
\end{tabular}
\end{center}
\caption{Comparison on different platforms.
}
\label{GPU comparisions}
\end{table}
\indent We also compare our method to state-of-the-art SGM designs on different platforms, as shown in Table \ref{GPU comparisions}. 
We run the OpenCV implementation\cite{opencv} on the embedded ARM CPU as the baseline and compute speed/power ratios (FPS/Watt) for each method. 
SGM\cite{chacon2013n} is the fastest GPU implementation on the KITTI leaderboard. We test SGM\cite{chacon2013n} on both high-end GPU (Nvidia Titan X) and low-power embedded GPU (Nvidia Jetson TX2) with the same parameters as the configuration C2 of FP-Stereo. We can see that SGM\cite{chacon2013n} runs at the fastest speed on a power-hungry GPU but the performance is degraded to 29 FPS when running on the low-power embedded GPU.
Given the same parameters as \cite{chacon2013n}, FP-Stereo-C2 achieves the same accuracy and executes at a fast speed with 161 FPS at the full KITTI resolution. Regarding energy efficiency, ours consumes the least energy and achieves the best speed/power ratio with 145x FPS/Watt. 
\vspace{+0.2cm}
\section{Conclusion}
We propose FP-Stereo for building high-performance stereo matching pipelines on FPGAs automatically to meet different user requirements in \textit{accuracy}, \textit{speed} and \textit{hardware cost}. Multiple methods are supported for each stage of the stereo matching pipeline and effective approaches are proposed to fully exploit the parallelism, efficiently utilize the resources and adequately ensure the usability. 
We analyze the impact of different design choices to provide informative guidance for users to select suitable implementations. The comparisons on different platforms demonstrate the superior performance of FP-Stereo compared to the state-of-the-art methods. FP-Stereo is open-source and available at https://eeweiz.home.ece.ust.hk/.

\vspace{+0.3cm}
\section{Acknowledgements}
The work was supported by the GRF grant 16213118 from the Research Grants Council in Hong Kong.
\vspace{+0.4cm}

\balance
{
\bibliographystyle{IEEEtran}
\bibliography{egbib}
}

\end{document}